\pdfoutput=1

\documentclass[11pt]{article}

\usepackage[final]{acl}

\usepackage{times}
\usepackage{amsmath}
\usepackage{latexsym}
\usepackage{amssymb}
\usepackage{graphicx}
\usepackage{float}
\usepackage{pifont}
\usepackage{subfig}
\usepackage{colortbl}  
\usepackage{xcolor}
\usepackage{array}   

\usepackage[T1]{fontenc}

\usepackage[utf8]{inputenc}

\usepackage{microtype}
\usepackage{multirow}
\usepackage{amsmath}
\usepackage{bm}
\usepackage{booktabs}
\usepackage{makecell}
\usepackage{lipsum}

\usepackage{xcolor}
\newcommand{\xwnote}[1]{\textbf{\color{red}#1}}

\title{What Would Happen Next? Predicting Consequences \\ from An Event Causality Graph}

\author{Chuanhong Zhan \textsuperscript{1} $\dagger$ \hspace{0.5cm}  Wei Xiang \textsuperscript{2} $\dagger$ \hspace{0.5cm}  Chao Liang\textsuperscript{1} \hspace{0.5cm}  Bang Wang\textsuperscript{1} \thanks{\quad Corresponding author: Bang Wang} \\
	\textsuperscript{1} School of Electronic Information and Communications, \\
	Huazhong University of Science and Technology, Wuhan, China \\
    \texttt{\{zhanch, liangchao111, wangbang\}@hust.edu.cn} \\
	\textsuperscript{2} Faculty of Artificial Intelligence in Education, Central China Normal University. \\
	\texttt{xiangwei@ccnu.edu.cn}
}

\begin{document}
\maketitle
\begin{abstract}
Existing script event prediction task forcasts the subsequent event based on an event script chain. However, the evolution of historical events are more complicated in real world scenarios and the limited information provided by the event script chain also make it difficult to accurately predict subsequent events. This paper introduces a \textit{Causality Graph Event Prediction}(CGEP) task that forecasting consequential event based on an Event Causality Graph (ECG). We propose a \textit{Semantic Enhanced Distance-sensitive Graph Prompt Learning} (\textsf{SeDGPL}) Model for the CGEP task. In \textsf{SeDGPL}, (1) we design a \textit{Distance-sensitive Graph Linearization (DsGL) } module to reformulate the ECG into a graph prompt template as the input of a PLM; (2) propose an \textit{Event-Enriched Causality Encoding (EeCE)} module to integrate both event contextual semantic and graph schema information; (3) propose a \textit{Semantic Contrast Event Prediction (ScEP)} module to enhance the event representation among numerous candidate events and predict consequential event following prompt learning paradigm. 
Experiment results validate our argument our proposed \textsf{SeDGPL} model outperforms the advanced competitors for the CGEP task.\footnote{We released the code at: \url{https://github.com/zhanchuanhong/SeDGPL}.}
\let\thefootnote\relax\footnotetext{$\dagger$ Authors marked with this symbol ($\dagger$) are co-authors.}
\end{abstract}

\begin{figure}[h]
	\centerline{\includegraphics[width=0.9\linewidth]{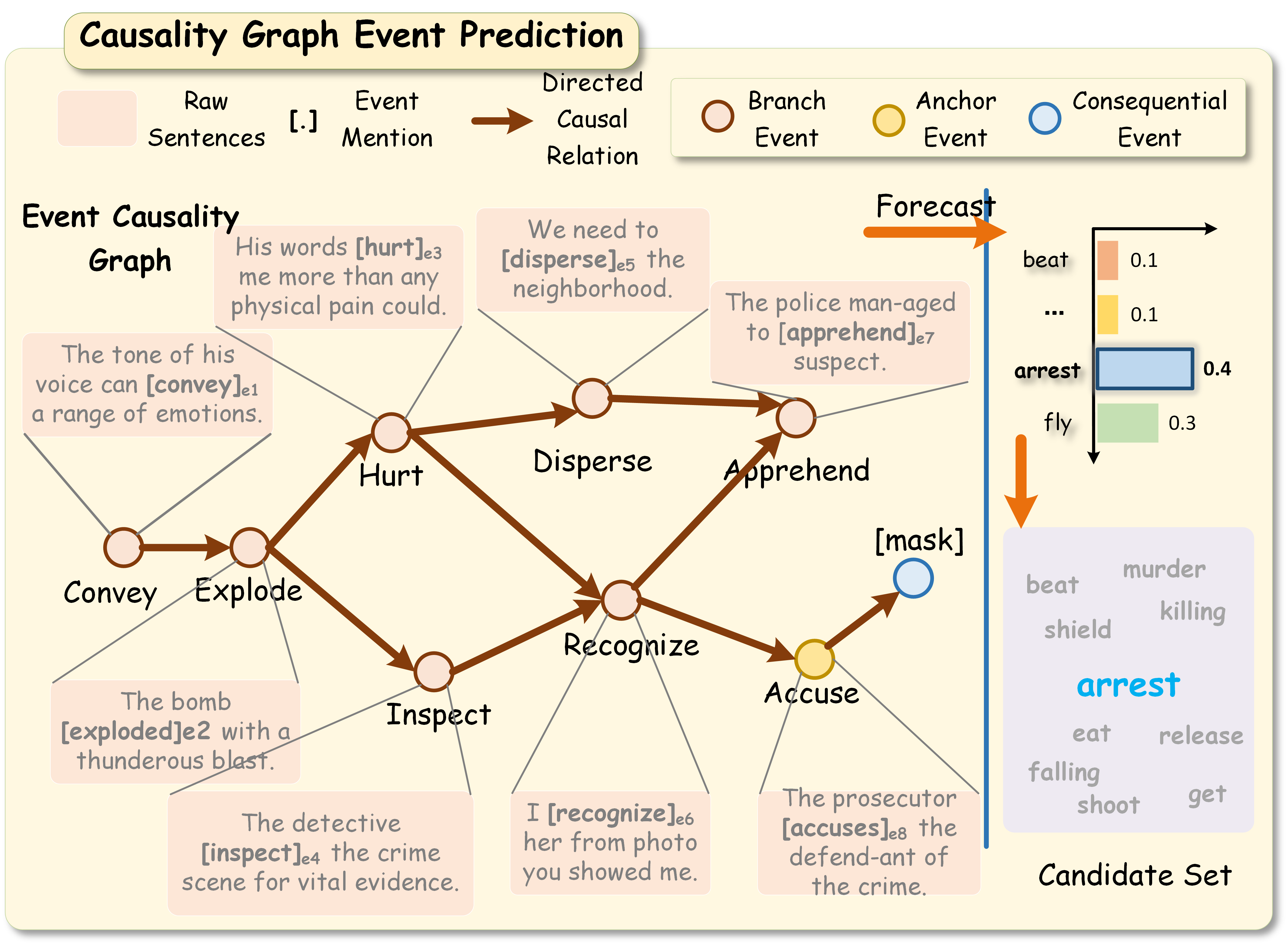}}
	\caption{Illustration of the Causality Graph Event Prediction task, forecasting consequential event based on an Event Causality Graph.}
	\label{Fig:Instance}
\end{figure}

\section{Introduction}
Event prediction aims to forecast the consequential event that are most likely to happen next, based on historical events and their relationships.
It has significant applications in many scenarios and industries, such as dialogue systems~\cite{Chen:et.al:2017:ACM}, discourse understanding~\cite{Lee:et.al:2019:ACL}, and story generation~\cite{Chaturvedi:et.al:2017:EMNLP}.
Existing script event prediction task~\cite{Wang:et.al:2023:IEEE} predicts the subsequent event given a sequence of events, named event script chain. However, we argue that the evolution of historical events are more complicated than a script event chain in real world scenarios. Besides, the limited information provided by event chains also make it difficult to accurately predict subsequent events.

\par
Motivated from such considerations, this paper introduces a \textit{Causality Graph Event Prediction}(CGEP) task that forecasting consequential event based on an Event Causality Graph (ECG).
As illustrated in Fig.~\ref{Fig:Instance}, the CGEP task is to select the most likely consequential event from candidate set based on an input ECG and an selected anchor event.
Instead of using event script chain for subsequent event prediction, we model the connection between events by an ECG, which can better reveal the evolution of  historical events. Besides, an ECG may have more than one consequential event that are likely to happen next. As such, we predict a consequential event for each tail node event (i.e. the anchor event) in an ECG, to achieve a more comprehensive understanding of events' evolution.

\par
Traditional event prediction methods either encode the contextual semantic of events~\cite{Du:et.al:2022:arXiv} or model the information of graph structure~\cite{Shirai:er.al:2023:WWW} for event  forcasting. The recently emerged prompt learning paradigm, based on pre-trained language model (PLM), exhibits outstanding ability in logical reasoning and has been applied in many natural language processing tasks~\cite{Xiang:et.al:2022:COLING}. However, most PLMs take text sequences as input and struggle to process graph-structured inputs. In this paper, we use graph prompt learning paradigm to linearize the input ECG, so as to utilize the encyclopedia-like knowledge embedded in a PLM for prediction.

\par
Besides, some studies obtain common sense knowledge from external knowledge bases to augment event prediction~\cite{Li:et.al:2018:arXiv}. This again validates our argument that the event chain input contains insufficiency information of historical event. By contrast our ECG input itself has included abundant historical events and causalities information. To this end, we enrich the event representation by integrating event contextual semantic and graph schema information from the input ECGs. Furthermore, we select the consequential event from a significantly larger candidate set than that of event script prediction task. And a semantic contrastive learning is used to enhance the event representation among numerous candidate events.

\par
In this paper, we introduce a CGEP task to forecast consequential event based on an ECG, and propose a \textit{Semantic Enhanced Distance-sensitive Graph Prompt Learning} (\textsf{SeDGPL}) Model for the CGEP task. The \textsf{SeDGPL} model contains three modules: (1) The \textit{Distance-sensitive Graph Linearization (DsGL) }module reformualtes the ECG into a graph prompt template as the input of a PLM; (2) The \textit{Event-Enriched Causality Encoding (EeCE)} module enriches the event representation by integrating both event contextual semantic and graph schema information; (3) The \textit{Semantic Contrast Event Prediction (ScEP)} module enhances the event representation among numerous candidate events and predicts consequential event following prompt learning paradigm.

\par
We construct two CGEP datasets based on existing event causality corpus MAVEN-ERE and Event StoryLine Corpus (ESC). Experiment results validate our argument that predicting events based on ECG is more reasonable than that based on event script chain, and our proposed \textsf{SeDGPL} model outperforms the advanced competitors for the task.

\section{Related Work}
\subsection{Script Event Prediction}
Script Event Prediction focuses on predicting future events based on a narrative event chain with shared entities.
Previous studies~\citep{Zhou:et.al:2022:IJIS,Wang:et.al:2021:WWW,Huang:et.al:2021:ACM} employ word2vec to encode the events, and predict subsequent events based on the similarity between candidate events and script events. 
With respect to temporal ordering, ~\citet{Pichotta:et.al:2016:AAAI,Wang:et.al:2017:EMNLP} employ Long Short-Term Memory (LSTM) to model the temporal dependencies between events. Contemporary event modeling methods utilize the Pre-trained Language Models, e.g. BERT~\citep{Devlin:et.al:arXiv:2018} and RoBERTa~\citep{Liu:et.al:2019:arXiv}. However, these models lack discourse-awareness as they are trained using Masked Language Modeling, which does not effectively capture the causal and temporal relations between multi-hop events. To address this problem, some research~\citep{Li:et.al:2018:arXiv,Zheng:et.al:2020:COLING} also explore specific event graphs as external knowledge base to assist event prediction. For example, ~\citet{Wang:et.al:2022:ACM} proposes a novel Retrieval-Enhanced Temporal Event forecasting framework, which dynamically retrieves high-quality sub-graphs based on the corresponding entities.

\subsection{Event Graph Reasoning}
Event Graph Reasoning aims to leverage the structure and connections within the graph to identify new patterns \citep{Roy:et.al:2024:AAAI} that do not explicitly exist in the event graph. Depending on the goal of reasoning, the task can be further categorized into relational reasoning~\citep{Huang:et.al:2024:Nips} and event prediction~\citep{li:et.al:2021:EMNLP}. For relation reasoning, ~\citet{Tang:et.al:2023:IEEE} adopts different event attributes to learn the semantic representations of events, and reasons event relation based on their similarities. ~\citet{Tang:et.al:2021:Springer} combines LSTM and attention mechanisms to dynamically generate event sequence representations, thereby predicting event relationships. For event prediction, prior studies~\cite{Du.Li:et.al:2021:ACL,Du.Xinya:et.al:2022:NAACL} perform subgraph matching between instance graph and schema graph to identify subsequent events. However, such methods predict event types rather than the events themselves. Moreover, ~\citet{Li:et.al:2023:Comp.sys,Islam:et.al:2024:arxiv} predict potential events for the next timestamp by dividing the event graph into a series of subgraphs based on event timestamps. 


\begin{figure*}[h]
	\centerline{\includegraphics[width=0.9\linewidth]{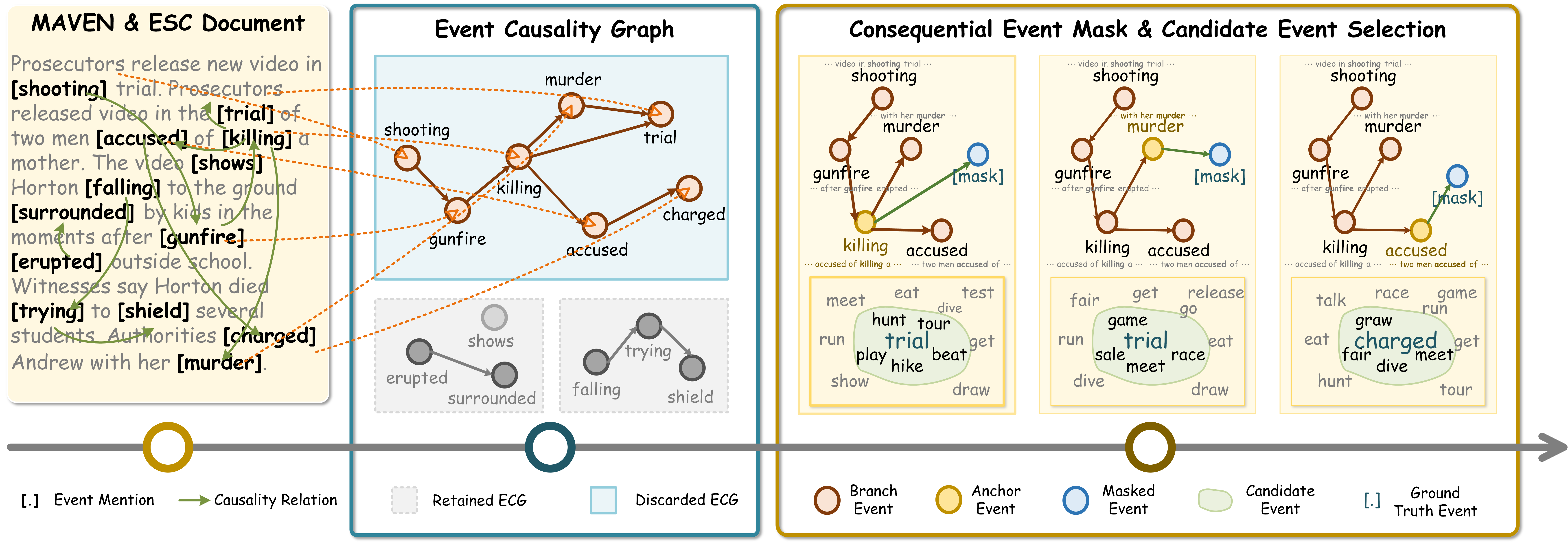}}
	\caption{Data Processing Flowchart: The data processing involves transforming an original ECG into multiple data instances, with each instance specifically predicting a single consequential event.}
	\label{Fig:dataprocess}
\end{figure*}

\section{Causality Graph Event Prediction}
\subsection{Task Definition}
We define the Causality Graph Event Prediction (CGEP) task as predicting the most likely consequential events that will occur next in an event causality graph (ECG).
As illustrated in Figure~\ref{Fig:Instance}, the ECG is a directed graph consisting of some past events as nodes and the causal relations between them as directed edges, denoted by $\mathcal{G} (\mathcal{E}, \mathcal{R})$. Where an event node $e_i \in \mathcal{E}$ contains the event mention word(s) $Em_i$ and the raw sentence $S_i$ it belongs to; A causality edge $r_{ij} \in \mathcal{R}$ is a directed causal relation from the event node $e_i$ to the event node $e_j$, indicating that $e_i$ causes $e_j$ (i.e. $e_i \rightarrow e_j$).
Each tail node in an ECG, which has no edge starting from to any other event node, is used as the anchor event $e_t \in \mathcal{E}$ for next event prediction.
The objective of CGEP task is to select the most likely consequential events $e_c$ from the candidate event set $\mathcal{E}_c$ for an anchor event node $e_t$ in an ECG.

\subsection{Datasets Construction}
We construct two CGEP datasets based on the public event causality dataset MAVEN-ERE~\cite{Wang:et.al:2022:arXiv} and EventStoryLine Corpus (ESC)~\cite{Caselli:et.al:2017:ACL}, annotating event mentions and directed causal relations between events within documents. Figure~\ref{Fig:dataprocess} illustrates the process of CGEP dataset construction.

\par
We first construct ECGs based on the annotations in each document from ESC and MAVEN-ERE datasets, using the annotated events as nodes and the annotated directed causal relation between events as edges.
Note that multiple disconnected ECGs may be constructed from a single document, and only \textit{weakly connected graphs}\xwnote{\footnote{A graph is considered weakly connected if every pair of vertices in the graph is connected by a path, regardless of the direction of the edges.}} with more than 4 event nodes are retained to ensure a complete event causality graph structure for event prediction.
We then mask one of the tail event node in an ECG as a CGEP instance, where the masked event is the consequential event $e_c$ to be predicted and its cause event is the anchor event $e_t$. In case that the masked event is caused by multiple events or an anchor event causes multiple effect events, it is further divided into multiple CGEP instances to ensure that each instance has an unique anchor event and ground truth consequential event.

\begin{table*}[]
	\centering
	\scalebox{1.0}{
		\renewcommand{\arraystretch}{1.2}
		\begin{tabular}{c|c|c|c|c|c|c}
			\toprule
			\midrule
			\rowcolor[HTML]{ECF4FF}
			Datasets    & Docs & ECGs & Avg.Nodes & Avg.Edges & Instances & CandiSet \\ \midrule
			CGEP-MAVEN & 3,015 & 5,308 & 8.4       & 12.9      & 12,200     & 512      \\
			CGEP-ESC   & 243   & 363   & 11        & 24.9      & 1,191      & 256      \\ \midrule
			\bottomrule
	\end{tabular}}
	\caption{Statistics of our CGEP-MAVEN and CGEP-ESC datasets.}
	\label{Tab:ProcessedStatistic}
\end{table*}

\par
For each CGEP instance, we randomly select a large number of tail node events from all other ECGs in the dataset as negative samples to construct a candidate set of consequential events $\mathcal{E}_c$. The ground truth event $e_c$ is the one that has been masked aforementioned.
Considering that the ground truth event mention may also appears in the sentence of other event nodes, that is the sentence it belongs to contains multiple event mentions, we replace them by a PLM-specific token [PAD] to prevent answer leakage.
Finally, we construct two CGEP dataset CGEP-MAVEN and CGEP-ESC\footnote{Datasets will be released publicly after the anonymous review.} for the CGEP task, in which each instance contains an event causality graph $\mathcal{G} (\mathcal{E}, \mathcal{R})$, an anchor event $e_t$, a candidate event set $\mathcal{E}_c$, and a ground truth consequential event $e_c$.

\par
Considering the varying instance sizes of the CGEP-MAVEN and CGEP-ESC datasets, the number of candidate sets for consequential events is randomly selected to be 512 and 256 , respectively.
Table~\ref{Tab:ProcessedStatistic} summarizes the statistics of our constructed CGEP-MAVEN and CGEP-ESC datasets.

\section{Methodology}

\begin{figure*}[h]
	\centerline{\includegraphics[width=0.9\textwidth]{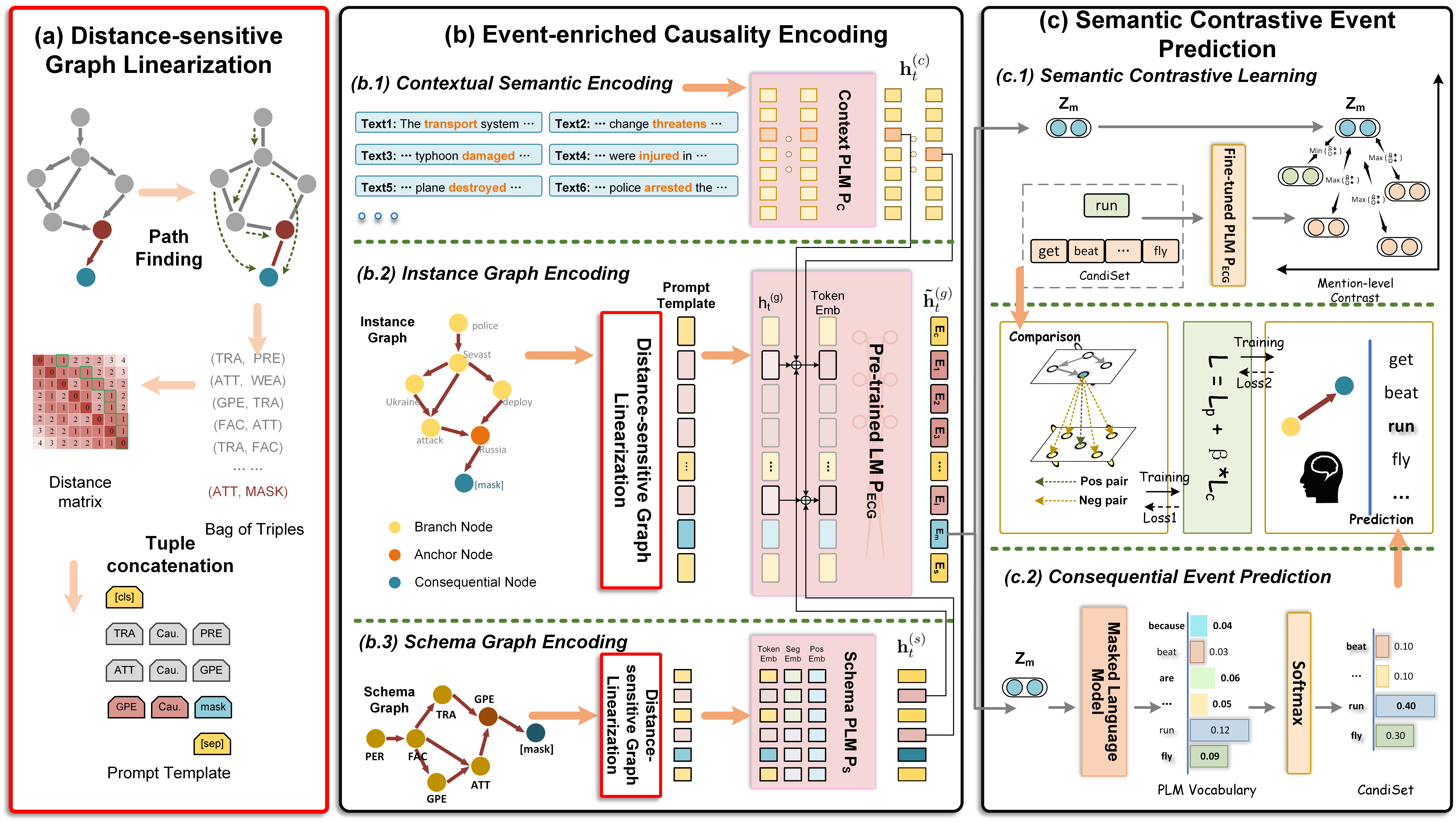}}
	\caption{The \textsf{SeDGPL} model consists of three modules: (1) \textit{Distance-sensitive Graph Linearization} (DsGL); (2) \textit{Event-Enriched Causality Encoding} (EeCE); (3) \textit{Semantic Contrast Event Prediction} (ScEP).}
	\label{Fig:Framework}
	\vspace{-10pt}
\end{figure*}

We propose a \textit{Semantic Enhanced Distance-sensitive Graph Prompt Learning Model} (\textsf{SeDGPL}) for causality graph event prediction. As illustrated in Figure~\ref{Fig:Framework}, the \textsf{SeDGPL} contains three modules:
(1) \textit{Distance-sensitive Graph Linearization} (DsGL); (2) \textit{Event-Enriched Causality Encoding} (EeCE); (3) \textit{Semantic Contrast Event Prediction} (ScEP).

\subsection{Distance-sensitive Graph Linearization}
The DsGL module is to reformulate the Event Causality Graph (ECG) of an input CGEP instance into a graph prompt template $\mathcal{T(G})$, as the input of a Pre-trained Language Model (PLM). As illustrated in Figure~\ref{Fig:Framework} (a), the graph prompt template $\mathcal{T(G})$ is a concatenation of some event causality triple templates $\mathcal{T}_n$ and a simple prompt template $\mathcal{T}_m$, represented as follows:
\begin{align} \label{Eq: Template}
	\mathcal{T(G}) = {\small \texttt{[C]}}, \mathcal{T}_1,   {\small \texttt{[S]}},  ... \   \mathcal{T}_n,   {\small \texttt{[S]}},  \mathcal{T}_m,  {\small \texttt{[S]}},
\end{align}
where $\small{\texttt{[C]}}$ and $\small{\texttt{[S]}}$ are the PLM-specific tokens $\small{\texttt{[CLS]}}$ and $\small{\texttt{[SEP]}}$, respectively, indicating the beginning and ending of an input sequence. Additionally, $\small{\texttt{[S]}}$ is also used to mark the boundary between each triple templates and the prompt template.

\par
Given an ECG $\mathcal{G}$ with $n$ directed causality edges, we can first obtain $n$ event causality triples $T_r^{(n)}=(e_i, r_{ij}, e_j)$, each containing a cause event $e_i$, an effect event $e_j$ and a directed causal relation $r_{ij}$ from $e_i$ to $e_j$. The template $\mathcal{T}_n$ for each event causality triple is formulated by concatenating of both the cause and the effect event mentions with an inserted conjunction word \textit{causes}:
\begin{align}
	\mathcal{T}_n = Em_i \ \ causes \ \ Em_j,
\end{align}
where $Em_i$ and $Em_j$  are the event mentions of cause event $e_i$ and effect event $e_j$, respectively.

\par
We argue that the closer an event causality triple $T_r^{(n)}$ is to the anchor event $e_t$, the stronger its connection to the anchor event, and it can provide more critical information for consequential event prediction.
To this end, we order the event causality triples based on their distance to the anchor event. The distance of an event causality triple $T_r^{(n)}=(e_i, r_{ij}, e_j)$ to the anchor event $e_t$ is computed by the number of edges on the shortest undirected path from its cause event $e_i$ to the anchor event $e_t$, as follows:
\begin{align}
	d_n (e_i, e_t) = \textrm{min}_{\ p \in P (e_i, e_t)} \  |p| ,
\end{align}
where $P(e_i, e_t)$ is the set of all undirected paths from the cause event $e_i$ to the anchor event $e_t$, and $|p|$ is the number of edges on the path $p$.

\par
We arrange the event causality triple templates $\mathcal{T}_n$ in decreasing order of their distance to the anchor event $e_t$. As in Equation~\ref{Eq: Template}, the distances are ordered such that $d_1 \geq d_2 \geq \ ...\  \geq d_n$, indicating that $\mathcal{T}_n$ is closest to the anchor event and $\mathcal{T}_1$ is the farthest one.
At the end of graph prompt template $\mathcal{T(G})$, we design and concatenate a simple prompt template $\mathcal{T}_m$ for event prediction:
\begin{align}
	\mathcal{T}_m = Em_t \ \ causes \ \ \texttt{[MASK]},
\end{align}
where $Em_t$ is the event mention of anchor event $e_t$ and the PLM-specific token $\texttt{[MASK]}$ is used to predict consequential event.

\subsection{Event-Enriched Causality Encoding}
To enrich the event representation for causality encoding, we propose an EeCE module that integrates both event contextual semantic and graph schema information into the ECG representation.
After graph linearization, we input each graph prompt template $\mathcal{T}(\mathcal{G})$ into a Pre-trained Language Model (PLM) for ECG encoding, denoted as $\mathcal{P}_{\textrm{ECG}}$.
As illustrated in Figure~\ref{Fig:Framework} (b), the input representation of PLM is constructed by summing the corresponding token embedding $\mathbf{h}_t^{(g)}$, the segment embedding $\mathbf{h}_s^{(g)}$, and the position embedding $\mathbf{h}_p^{(g)}$:
\begin{align}
	\mathbf{h}^{(g)} = \mathbf{h}_t^{(g)} + \mathbf{h}_s^{(g)} + \mathbf{h}_p^{(g)}.
\end{align}

\par
For contextual semantic encoding, we input the raw sentence $S_i$ of each event into another PLM $\mathcal{P}_c$ to obtain its contextual representation $\mathbf{h}^{(c)}$, as illustrated in Figure~\ref{Fig:Framework} (b.1).
For schema information encoding, we first construct an event schema graph by replacing each event node in an ECG with its corresponding annotated event type, like ~\citep{Zhuang:et.al:2023:Springer,Groz:et.al:2021:arxiv}, and etc.
After the same graph linearization operation, we input each schema graph template into another PLM $\mathcal{P}_s$ to obtain the event's schema representation $\mathbf{h}^{(s)}$, as illustrated in Figure~\ref{Fig:Framework} (b.3).
We note that only the token embeddings of event's contextual representation $\mathbf{h}_t^{(c)}$ and schema representation $\mathbf{h}_t^{(s)}$ are used for next enrichment fusion. The segment embedding $\mathbf{h}_s^{(g)}$ and position embedding $\mathbf{h}_p^{(g)}$ of ECG encoding, which contain graph structure information, are directly used without fusion.

\par
To fuse the features of event's contextual semantic and schema information into the ECG representation, we use the fusion gate to integrate their event's representations $\mathbf{h}_t^{(c)}$ and $\mathbf{h}_t^{(s)}$ into the event's representation of ECG $\mathbf{h}_t^{(g)}$.
Specifically, we first use a fusion gate to integrate the contextual representation $\mathbf{h}_t^{(c)}$ schema representation $\mathbf{h}_t^{(s)}$, and output $\mathbf{{h}}_t^{(r)} \in \mathbb{R}^{d_h}$ as the event enrichment vector. The transition functions are:
\begin{align} \label{}
	\mathbf{g}_{r} & = \textrm{sigmoid}(\mathbf{W}_{r} \mathbf{h}_t^{(c)} + \mathbf{U}_{r} \mathbf{h}_t^{(s)}), \\
	\mathbf{{h}}_t^{(r)} & = \mathbf{g}_{r} \odot \mathbf{h}_t^{(c)} + (1 - \mathbf{g}_{r}) \odot \mathbf{h}_t^{(s)},
\end{align}
where $\mathbf{W}_{r} \in \mathbb{R}^{d_h \times d_h}$, $\mathbf{U}_{r} \in \mathbb{R}^{d_h \times d_h}$ are learnable parameters and $\odot$ donates the element-wise product of vectors.

\par
We next use another fusion gate to integrate the event enrichment vector $\mathbf{{h}}_t^{(r)}  \in \mathbb{R}^{d_h}$ into the token embeddings of event's representation in ECG $\mathbf{h}_t^{(g)}$. The transition functions are:
\begin{align} \label{}
	\mathbf{g}_e & = \textrm{sigmoid}(\mathbf{W}_e \mathbf{h}_t^{(g)} + \mathbf{U}_e \mathbf{{h}}_t^{(r)}), \\
	\mathbf{\tilde{h}}_t^{(g)} & = \mathbf{g}_e \odot \mathbf{h}_t^{(g)} + (1 - \mathbf{g}_e) \odot \mathbf{{h}}_t^{(r)},
\end{align}
where $\mathbf{W}_e \in \mathbb{R}^{d_h \times d_h}$, $\mathbf{U}_e \in \mathbb{R}^{d_h \times d_h}$ are learnable parameters.
With the fusion gate, we enrich the event's representation in ECGs by integrating both event's contextual semantic and schema information features.
Note that only the representations of event mention in ECGs are fused, the other tokens in graph prompt template $\mathcal{T(G})$, such as \textit{causes}, ${\texttt{[CLS]}}$, ${\texttt{[SEP]}}$, ${\texttt{[MASK]}}$, and etc., are originally encoded by the ECG encoding PLM $\mathcal{P}_{\textrm{ECG}}$.

\par
Finally, the PLM $\mathcal{P}_{\textrm{ECG}}$ outputs a hidden state vector $\mathbf{z}$ for each input token in the graph prompt template $\mathcal{T}(\mathcal{G})$, using the fused event's token embeddings as input representations.

\begin{table*}[ht]
	\centering
	\resizebox{\linewidth}{!}{
		\renewcommand{\arraystretch}{1.3}
		\begin{tabular}{c|cccccc|cccccc}
			\toprule
			\midrule &\multicolumn{6}{c|}{\cellcolor[HTML]{CBCEFB}\textbf{CGEP-MAVEN}} & \multicolumn{6}{c}{\cellcolor[HTML]{CBCEFB}\textbf{CGEP-ESC}} \\ \cmidrule{2-13} \multirow{-2}{*}{\textbf{Model}} & \cellcolor[HTML]{ECF4FF}MRR & \cellcolor[HTML]{ECF4FF}Hit@1 & \cellcolor[HTML]{ECF4FF}Hit@3 & \cellcolor[HTML]{ECF4FF}Hit@10 & \cellcolor[HTML]{ECF4FF}Hit@20 & \cellcolor[HTML]{ECF4FF}Hit@50 & \cellcolor[HTML]{ECF4FF}MRR & \cellcolor[HTML]{ECF4FF}Hit@1 & \cellcolor[HTML]{ECF4FF}Hit@3 & \cellcolor[HTML]{ECF4FF}Hit@10 & \cellcolor[HTML]{ECF4FF}Hit@20 & \cellcolor[HTML]{ECF4FF}Hit@50 \\ \midrule
			\textsf{CSProm-KG} &
			\cellcolor[HTML]{FAEBD7}22.3 & \cellcolor[HTML]{FAEBD7}18.1 & \cellcolor[HTML]{FAEBD7}23.2 & \cellcolor[HTML]{FAEBD7}31.0 & \cellcolor[HTML]{FAEBD7}38.4 & \cellcolor[HTML]{FAEBD7}50.7 & 14.2 & \cellcolor[HTML]{FAEBD7}11.9 & 11.3 & \cellcolor[HTML]{FAEBD7}21.0 & \cellcolor[HTML]{FAEBD7}25.6 & 34.6 \\
			\textsf{SimKG} &
			9.3 & 4.5 & 9.2 & 18.0 & 25.3 & 35.0 & \cellcolor[HTML]{FAEBD7}14.9 & 10.3 & \cellcolor[HTML]{FAEBD7}13.5 & 18.4 & 22.3 & 34.0 \\
			\textsf{BARTbase} &
			\cellcolor[HTML]{FFDEAD}24.7 & \cellcolor[HTML]{FFDEAD}19.5 & \cellcolor[HTML]{FFDEAD}24.5 & \cellcolor[HTML]{FFDEAD}34.8 & \cellcolor[HTML]{FFDEAD}42.6 & \cellcolor[HTML]{FFDEAD}53.6 & \cellcolor[HTML]{FFDEAD}16.0 & \cellcolor[HTML]{FFDEAD}12.5 & \cellcolor[HTML]{FFDEAD}16.8 & \cellcolor[HTML]{FFDEAD}21.1 & \cellcolor[HTML]{FFDEAD}28.6 & \cellcolor[HTML]{FFDEAD}38.9 \\
			\textsf{MCPredictor} &
			18.1 & 13.0 & 18.4 & 27.3 & 32.0  & 43.2 & 9.7 & 8.4 & 10.9 & 17.4 & 22.2 & \cellcolor[HTML]{FAEBD7}37.5\\ \midrule
			\textsf{Llama3-7B} &
			9.6 & 5.0 & 11.1 & 20.2 & 24.5 & 26.6 & 6.7 & 1.1 & 8.9 & 20.2 & 26.3 & 29.2 \\
			\textsf{GPT-3.5-turbo} &
			14.6 & 8.1 & 17.1 & 28.1 & 33.3 & 39.5 & 10.1 & 4.9 & 11.4 & 20.5 & 25.2 & 31.5 \\ \midrule
			\textsf{SeDGPL} &
			\cellcolor[HTML]{FFDAB9}\textbf{27.9} & \cellcolor[HTML]{FFDAB9}\textbf{21.9} & \cellcolor[HTML]{FFDAB9}\textbf{28.9} & \cellcolor[HTML]{FFDAB9}\textbf{40.8} & \cellcolor[HTML]{FFDAB9}\textbf{48.1} & \cellcolor[HTML]{FFDAB9}\textbf{57.9} & \cellcolor[HTML]{FFDAB9}\textbf{19.6} & \cellcolor[HTML]{FFDAB9}\textbf{15.2} & \cellcolor[HTML]{FFDAB9}\textbf{18.1} & \cellcolor[HTML]{FFDAB9}\textbf{22.3} & \cellcolor[HTML]{FFDAB9}\textbf{29.9} & \cellcolor[HTML]{FFDAB9}\textbf{41.9} \\ \midrule
			\bottomrule
	\end{tabular}}
	\caption{Overall results of Causality Graph Event Prediction on the CGEP-MAVEN and CGEP-ESC datasets.}
	\label{table:graph-main}
\end{table*}

\begin{table}[ht]
	\centering
	\resizebox{\linewidth}{!}{
		\renewcommand{\arraystretch}{1.1}
		\begin{tabular}{c|ccccc}
			\toprule
			\midrule &
			\multicolumn{5}{c}{\cellcolor[HTML]{CBCEFB}\textbf{CGEP-MAVEN}}  \\ \cmidrule{2-6}
			\multirow{-2}{*}{\textbf{Model}} &
			\cellcolor[HTML]{ECF4FF}MRR & \cellcolor[HTML]{ECF4FF}Hit@1 & \cellcolor[HTML]{ECF4FF}Hit@3 & \cellcolor[HTML]{ECF4FF}Hit@10 & \cellcolor[HTML]{ECF4FF}Hit@50 \\
			\textsf{CSProm-KG} &
			7.1 ($\downarrow$15.2) & \cellcolor[HTML]{FAEBD7}4.8 ($\downarrow$13.3) & 6.4($\downarrow$16.8) & 10.6($\downarrow$20.4) & 22.4($\downarrow$28.3)  \\
			\textsf{SimKG} &
			5.0($\downarrow$4.3) & 2.2($\downarrow$2.3) & 4.3($\downarrow$4.9) & 8.5($\downarrow$9.5) & 25.7($\downarrow$9.3)  \\
			\textsf{BARTbase} &
			\cellcolor[HTML]{FFDEAD}11.8($\downarrow$12.9) & \cellcolor[HTML]{FFDEAD}8.2($\downarrow$11.3) & \cellcolor[HTML]{FFDEAD}11.2($\downarrow$13.3) & \cellcolor[HTML]{FFDEAD}16.6($\downarrow$18.2)  & \cellcolor[HTML]{FFDEAD}34.2($\downarrow$19.4) \\
			\textsf{MCPredictor} &
			\cellcolor[HTML]{FAEBD7}7.3($\downarrow$10.8) & 3.6($\downarrow$9.4) & \cellcolor[HTML]{FAEBD7}7.3($\downarrow$14.5) & \cellcolor[HTML]{FAEBD7}14.8($\downarrow$19.7) & \cellcolor[HTML]{FAEBD7}29.4($\downarrow$13.8) \\ \midrule
			\textsf{SeDGPL} &
			\cellcolor[HTML]{FFDAB9}\textbf{16.0}($\downarrow$11.9) & \cellcolor[HTML]{FFDAB9}\textbf{12.4}($\downarrow$9.5) & \cellcolor[HTML]{FFDAB9}\textbf{15.4}($\downarrow$13.5) & \cellcolor[HTML]{FFDAB9}\textbf{23.0}($\downarrow$17.8) & \cellcolor[HTML]{FFDAB9}\textbf{39.4}($\downarrow$18.5) \\ \midrule
			\bottomrule
	\end{tabular}}
	\caption{Overall results of Script Event Prediction on CGEP-MAVEN dataset.}
	\label{table:script-main}
	\vspace{-10pt}
\end{table}

\subsection{Semantic Contrast Event Prediction}
Following the prompt learning paradigm~\cite{Xiang:et.al:2023:arXiv,Li:et.al:2023:ACL}, we use the hidden state vector of \texttt{[MASK]} token $\mathbf{z}_m$ for consequential event prediction.
To enhance the PLM's ability of understanding event semantic among numerous candidate events, we apply a kind of semantic contrastive learning to improve the \texttt{[MASK]} token presentation $\mathbf{z}_m$.

\par
\textbf{Semantic Contrastive Learning:} As illustrated in Figure~\ref{Fig:Framework} (c.1), we first obtain a representation vector $\mathbf{z}_c$ for each candidate event $e_c$ using the fine-tuned PLM $\mathcal{P}_{ECG}$.
Then, the hidden state of \texttt{[MASK]} token $\mathbf{z}_m$ is used as the anchor sample, and the candidate event representations $\mathbf{z}_c$ are used as contrastive samples, where the ground truth event is the positive sample $\mathbf{z}_c^+$ and the other candidate events are negative samples $\mathbf{z}_c^-$. We employ the Supervised contrastive loss~\citep{Khosla:et.al:2020:NeurIPS} to compute the semantic contrast loss, as follows:
\begin{equation}
	{L}_{c}=-\log  \dfrac{\exp(\mathbf{z}_m \cdot  \mathbf{z}_c^+/{\tau})} {\sum\limits_{c \in \mathcal{C}} \exp(\mathbf{z}_m \cdot  \mathbf{z}_c /{\tau})},
\end{equation}
where $\tau$ is a scalar temperature parameter and $\mathcal{C}$ is the candidate set containing the positive sample and negative samples.

\par
\textbf{Consequential Event Prediction:} As illustrated in Figure~\ref{Fig:Framework} (c.2), the PLM $\mathcal{P}_{\textrm{ECG}}$ estimates the probability of each word within its vocabulary $V$ for the hidden state of \texttt{[MASK]} token $\mathbf{z}_m$. We use the predicted probability of the event mention word $e_c$ in the event candidate set $\mathcal{E}_c$ as the ranking score, to form an event prediction list:
\begin{align} \label{}
	\mathcal{P} ({\texttt{[MASK]}} = e_c \in \mathcal{E}_c \ | \ \mathcal{T}(\mathcal{G})).
\end{align}
We employ the cross entropy loss to compute the event prediction loss, as follows:
\begin{align} \label{}
	{L}_{p} = -\frac{1}{K} \sum_{k=1}^{{K}}  \mathbf{y}^{(k)} \log(\mathbf{\hat{y}}^{(k)}) + \lambda \Vert \theta \Vert^2,
\end{align}
where $\mathbf{y}^{(k)}$ and $\mathbf{\hat{y}}^{(k)}$ are the gold label and predicted label of the $k$-th training instance respectively. $\lambda$ and $\theta$ are the regularization hyper-parameters. We use the AdamW optimizer~\citep{Loshchilov:et.al:2017:arXiv} with $L2$ regularization for model training.

\textbf{Training Strategy:}  The cost function of our \textsf{SeDGPL} is optimized as follows:
\begin{equation}
	{L} = {L}_p+\beta * {L}_c,
\end{equation}
where $\beta$ is a weight coefficient to balance the importance of the event prediction loss and semantic contrast loss.

\section{Experiment}
\subsection{Experiment Settings}
Our experiments are conducted using the constructed CGEP-MAVEN and CGEP-ESC datasets. Following the standard data splitting of the underlying ESC~\cite{Caselli:et.al:2017:ACL} corpus, we use the last two topics as development set and conduct 5-fold cross-validation on the remaining 20 topics.
The average results of each fold are adopted as performance metrics.
Since the underlying MAVEN-ERE corpus did not release the test set, following \cite{Tao:et.al:2023:ACL}, we use the original development set as our test set and sample 20\% of the data from the original training set to form the development set. 
\par
We adopt the MRR (Mean Reciprocal Rank) and Hit@n (Hit Rate at n) as the evaluation metrics. Details about experimental settings and evaluation metrics can be found in Appendix~\ref{App:ExpSet}.

%

\subsection{Competitors}
We replicate some advanced event prediction models to conduct causality graph event prediction as benchmarks, including methods in knowledge graph completion tasks (\textsf{CSProm-KG}~\cite{Chen:et.al:2023:ACL}, \textsf{SimKG}~\cite{Wang:et.al:2022:ACL}) and script event prediction (\textsf{BARTbase}~\cite{Zhu:et.al:2023:AAAI}, \textsf{MCPredictor}~\cite{Bai:et.al:2021:EMNLP}). Furthermore, we validate the effectiveness of large language models on the CGEP task, including \textsf{Llama3-7B}~\cite{Touvron:et.al:2023:arXiv} and \textsf{GPT-3.5-turbo}~\cite{Gao:et.at:2023:ACL-findings}. For more details about its specific implementation, please refer to the Appendix~\ref{App:Comp.} and Appendix~\ref{GPT}.
%

\begin{table*}[]
	\resizebox{\linewidth}{!}{
		\renewcommand{\arraystretch}{1.3}
		\begin{tabular}{c|cccccc|cccccc}
			\toprule
			\midrule & \multicolumn{6}{c|}{\cellcolor[HTML]{CBCEFB}\textbf{CGEP-MAVEN}} & \multicolumn{6}{c}{\cellcolor[HTML]{CBCEFB}\textbf{CGEP-ESC}} \\ \cmidrule{2-13}
			\multirow{-2}{*}{\textbf{Model}} &
			\cellcolor[HTML]{ECF4FF}MRR & \cellcolor[HTML]{ECF4FF}Hit@1 & \cellcolor[HTML]{ECF4FF}Hit@3 & \cellcolor[HTML]{ECF4FF}Hit@10 & \cellcolor[HTML]{ECF4FF}Hit@20 & \cellcolor[HTML]{ECF4FF}Hit@50 & \cellcolor[HTML]{ECF4FF}MRR & \cellcolor[HTML]{ECF4FF}Hit@1 & \cellcolor[HTML]{ECF4FF}Hit@3  & \cellcolor[HTML]{ECF4FF}Hit@10 & \cellcolor[HTML]{ECF4FF}Hit@20 & \cellcolor[HTML]{ECF4FF}Hit@50 \\ \midrule
			\textsf{SeDGPL w/o Dist.}
			& \cellcolor[HTML]{FFDEAD}26.4 & \cellcolor[HTML]{FFDEAD}20.4 & \cellcolor[HTML]{FFDEAD}26.2 & \cellcolor[HTML]{FFDEAD}39.2 & \cellcolor[HTML]{FFDEAD}47.0 & \cellcolor[HTML]{FFDEAD}57.2 & \cellcolor[HTML]{FAEBD7}13.9 & 7.8 & \cellcolor[HTML]{FFDEAD}15.6 & 18.8 & 23.9  & \cellcolor[HTML]{FAEBD7}37.8 \\
			\textsf{SeDGPL w/o Ctxt.}
			& 5.3 & 4.0 & 4.2 & 9.6 & 13.9 & 23.6 & 12.2 & \cellcolor[HTML]{FAEBD7}8.8 & 11.0 & 17.9 & 21.7 & 33.8 \\
			\textsf{SeDGPL w/o Schm.}
			& \cellcolor[HTML]{FAEBD7}22.0 & \cellcolor[HTML]{FAEBD7}17.0 & \cellcolor[HTML]{FAEBD7}21.9 & 31.5 & 40.9 & \cellcolor[HTML]{FAEBD7}54.3 & \cellcolor[HTML]{FFDEAD}15.6 & \cellcolor[HTML]{FFDEAD}11.5 & 12.4 & \cellcolor[HTML]{FFDEAD}20.4 & \cellcolor[HTML]{FAEBD7}24.3 & 37.4 \\
			\textsf{SeDGPL w/o Ctrst.}
			& 21.2 & 15.8 & 21.0 & \cellcolor[HTML]{FAEBD7}32.0 & \cellcolor[HTML]{FAEBD7}41.4 & 53.8 & 13.2 & 8.5 & \cellcolor[HTML]{FAEBD7}14.5 & \cellcolor[HTML]{FAEBD7}20.0 & \cellcolor[HTML]{FFDEAD}25.2 & \cellcolor[HTML]{FFDEAD}38.0 \\
			
			\midrule
			\textsf{Full SeDGPL}
			& \cellcolor[HTML]{FFDAB9}\textbf{27.9} & \cellcolor[HTML]{FFDAB9}\textbf{21.9} & \cellcolor[HTML]{FFDAB9}\textbf{28.9} & \cellcolor[HTML]{FFDAB9}\textbf{40.8} & \cellcolor[HTML]{FFDAB9}\textbf{48.1} & \cellcolor[HTML]{FFDAB9}\textbf{57.9} & \cellcolor[HTML]{FFDAB9}\textbf{19.6} & \cellcolor[HTML]{FFDAB9}\textbf{15.2} & \cellcolor[HTML]{FFDAB9}\textbf{18.1} & \cellcolor[HTML]{FFDAB9}\textbf{22.3} & \cellcolor[HTML]{FFDAB9}\textbf{29.9} & \cellcolor[HTML]{FFDAB9}\textbf{41.9} \\ \midrule
			\bottomrule
	\end{tabular}}
	\caption{Experiment results of ablation study on both CGEP-MAVEN corpus and CGEP-ESC corpus.}
	\label{table:ablation}
\end{table*}

\subsection{Overall Results}
Table~\ref{table:graph-main} compares the overall performance between our \textsf{SeDGPL} and the competitors on both CGEP-MAVEN and CGEP-ESC datasets.

\par
We can observe that our \textsf{SeDGPL} has achieved significant performance improvements overall competitors in terms of much higher MRR and Hit@n. We attribute its outstanding performance to two main factors:
1) The transformation of the event causality graph into an ordered triple sequence for graph prompt learning, which enables our \textsf{SeDGPL} to effectively leverage both the structure information of event causality graph and the encyclopedic knowledge in a PLM for event prediction;
2) The enrichment of event representation through contextual semantic and schema information fusion encoding.
Besides, We can also observe that the \textsf{BARTbase} outperforms the other competitors in Table~\ref{table:graph-main}.
This might be attributed to the fine-tuning of a pre-trained language model in advance using an event-centric pre-training objective, which injects event-level knowledge into the PLM before making predictions.
3)  The performance of the \textsf{Llama3-7B} and \textsf{GPT-3.5-turbo} surpasses some models trained on the entire dataset, e.g. the \textsf{SimKG} model , indicating that large language models have great potential in understanding event relationships and reasoning event patterns.

\begin{figure}[h]
	\centerline{\includegraphics[width=0.9\columnwidth]{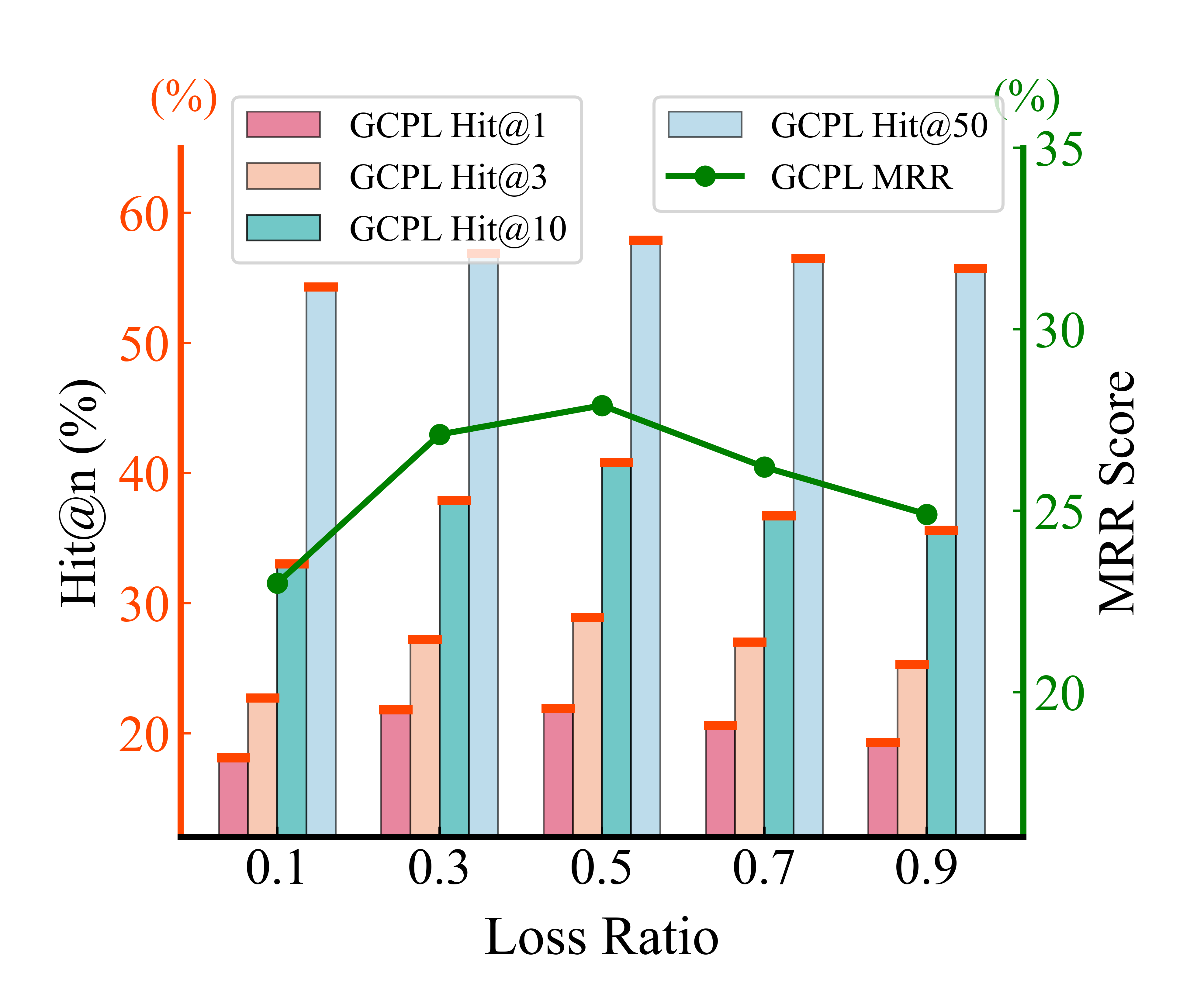}}
	\caption{Results on CGEP-MAVEN with different loss ratio $\beta$.}
	\label{Fig:lossratio}
\end{figure}

\begin{figure*}[h]
	\centerline{\includegraphics[width=0.9\textwidth]{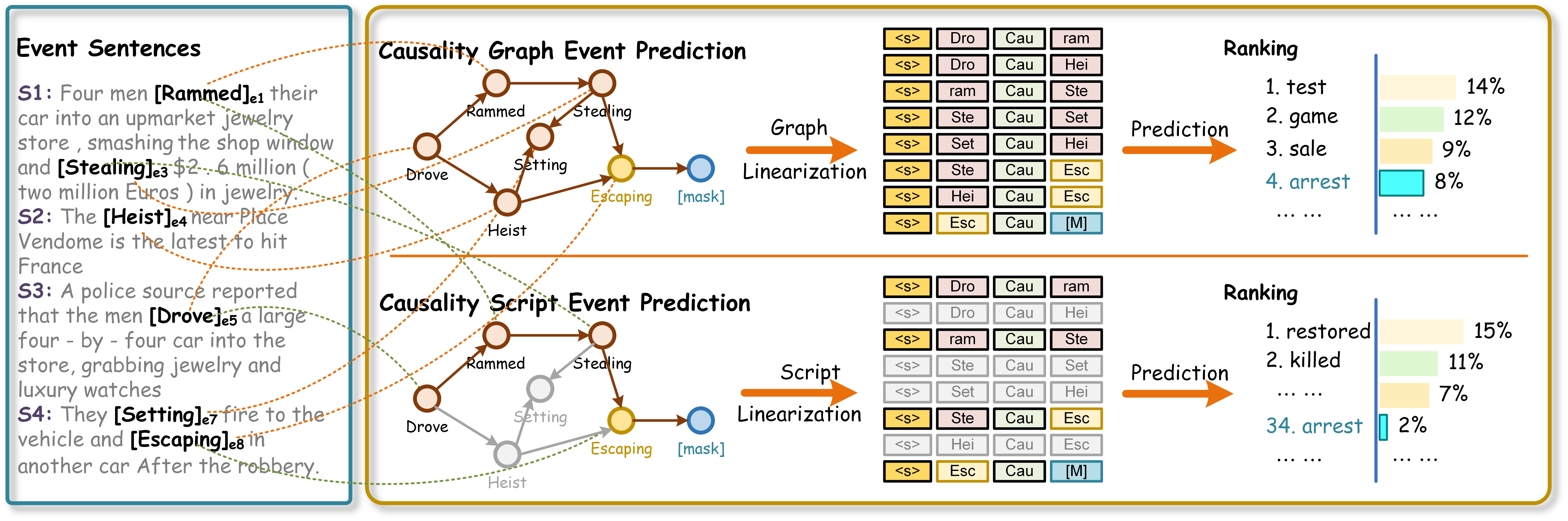}}
	\caption{A case of \textsf{SeDGPL} on causality graph event prediction and causality script event prediction tasks.}
	\label{Fig:case}
\end{figure*}

\par
To validate our argument that predicting consequential events based on event causality graph is more effective than predicting based on the event script chain, we also employ our \textsf{SeDGPL} and the competitors to conduct script event prediction for comparison, using the longest event chain in each event causality graph from CGEP-MAVEN dataset.\footnote{Considering the instance number of event chains in CGEP-ESC dataset, we only conduct script event prediction based on the CGEP-MAVEN dataset.}
Table~\ref{table:script-main} presents the performance of script event prediction between our \textsf{SeDGPL} and the competitors, as well as the performance variation compared with causality graph event prediction.
We can observe that the performance of event prediction suffers significantly due to the transformation of the causality graph input into the even chain input.
This is not unexpected. The event causality graph has a more complex structure than the script event chain, as it includes additional event nodes and causal connections, that can provide comprehensive prior knowledge for event prediction.
Besides, it can be observed that our \textsf{SeDGPL} also outperforms all competitors in script event prediction, again approving our design object.

\subsection{Ablation Study}

\paragraph{Module Ablation}To examine the effectiveness of different modules, we design the following ablation study: (1) \textsf{SeDGPL w/o Dist.} randomly orders the event causality triples without considering distance sensitivity; (2) \textsf{SeDGPL w/o Ctxt.} enriches event representation with only schema information, but without its contextual semantic; (3) \textsf{SeDGPL w/o Schm.} enriches event representation with only contextual semantic, but without its schema information; (4) \textsf{SeDGPL w/o Ctrst.} predicts consequential events without semantic contrastive learning. Table~\ref{table:ablation} presents the results of our module ablation study.

\par
The first observation is that neither the \textsf{SeDGPL w/o Ctxt.} and the \textsf{SeDGPL w/o Schm.} can outperform the \textsf{Full SeDGPL} model. This indicates that our fusion of both event contextual semantic and graph schema information is an effective approach to enrich event's representation learning for consequential event prediction. On the other hand, the \textsf{SeDGPL w/o Ctxt.} performs the worst among all ablation models. As it merely uses event mention words for representation learning, ignoring the event contextual semantic and existing linguistic ambiguities.
The second observation is that the \textsf{SeDGPL w/o Dist.} cannot outperform the \textsf{Full SeDGPL} model, even the performance gap is not obvious. This suggests that it is essential to order event causality triples based on distance sensitivity, as different triples in an event causality graph may have different importance for prediction consequential events.
We can also observe that the \textsf{SeDGPL w/o Ctrst.} cannot outperform the \textsf{Full SeDGPL} model, validating the effectiveness of contrastively learning the \texttt{[MASK]} token presentation $\mathbf{z}_m$ among numerous candidate events.

\paragraph{Hyper-parameter Ablation}
To further examine the impact of semantic contrastive learning module, we compare the performance of our \textsf{SeDGPL} against using different contrastive loss weight coefficient $\beta$ on the CGEP-MAVEN dataset, as plotted in Figure~\ref{Fig:lossratio}.
It can be observe that our \textsf{SeDGPL} achieves the best overall performance when the contrastive loss weight coefficient is set to 0.5. Yet it suffers from either a large or small value of the loss weight coefficient. Indeed, a small weight coefficient weakens the impact of semantic contrastive learning; By contrast, a large weight coefficient ignores the event prediction loss in back-propagation.

\section{Case study}
Figure~\ref{Fig:case} illustrates an example of \textsf{SeDGPL} applied to the causality graph event prediction (CGEP) task and the causality script event prediction (CSEP) task. For the CGEP task, \textsf{SeDGPL} linearizes the entire event graph into an event chain, comprehensively considering all the causality triples in the event graph. In contrast, for the CSEP task, \textsf{SeDGPL} extracts only a subset of the causality triples from the event graph to form the main event chain, disregarding the other nodes in the event graph, which undermines the structural information of the event graph. From Figure~\ref{Fig:case}, we observe that incorporating information beyond the main event chain can effectively aid the model in predicting subsequent events more accurately. For instance, in the CGEP task, given the causality triples \textit{(Drove, causes, Heist)} and \textit{(Heist, causes, Escaping)} as prior knowledge, our model can readily infer that the subsequent event following "Escaping" is "arrest". In contrast, for the CSEP task, the model only relies on the main event chain to judge the relationship between events. Therefore, the model can not effectively capture the causal relationships between events at different levels and the complex structure information in the event causality graph, leading to a decline in performance.

\section{Concluding Remarks}
In this paper, we argue that predicting consequential events based on the event causality graph is more effective than predicting based on the event script chain. To validate our argument, we propose the SeDGPL Model, a distance-sensitive graph prompt model that integrate both event contextual semantic and graph schema information, and conduct abundant experiments on both CGEP and SEP task. Experiment results validate our argument, and our proposed SeDGPL model outperforms the advanced competitors for the CGEP task.
\par
In our future work, we shall attempt to integrate other types of event relationships, e.g. temporal relations, to assist in event prediction.

\section{Limitation}
Due to the input length limitations of PLMs, we may have to discard some triplets during the linearization process, which could result in the loss of information beneficial for prediction.

\section{Acknowledge}
This work is supported in part by National Natural Science Foundation of China (Grant No:62172167). The computation is completed in the HPC Platform of Huazhong University of Science and Technology.

\section{Ethics Statement}
This paper has no particular ethic consideration.


\bibliography{anthology,custom}

\clearpage
\appendix

\section{Competitors} \label{App:Comp.}

\paragraph{\textsf{CSProm-KG}}We first linearize instance graphs and input them into the PLM, rather than modeling based on triplets as in the original model. And then we generate the conditional soft prompts based on the node embedding $E_h$ and relation embedding $E_r$, obtained by the embedding layer of RoBERTa. Then the conditional soft prompts will be concatenated with the text embedding and input into the PLM. Notice due to the specificity of the task, we cannot employ the KGC model ConvE for prediction. Consequently, we input the anchor node embedding into an MLP classification layer to obtain the final prediction.


\paragraph{\textsf{SimKG}}Similar to our proposed model, we initially transform the event instance graph to a linearized input sequence. To fully consider the impact of the candidate event set on the model, we also incorporate the candidate event set into contrastive learning, referring to them as \textit{candidate event negatives}. Since the consequence node to be predicted must be the 1-hop neighbors of the anchor node, we set the \textit{Re-Ranking} factors $\alpha$ to 0. Finally, we combine four types of contrastive losses, referred to as \textit{Candi-event Negatives}, \textit{In-batch Negatives}, \textit{Pre-batch Negatives}, and \textit{Self-Negativest}, and utilize back-propagation to update the model parameters.

\paragraph{\textsf{BARTbase}}We first populate the event instance graph according to the template described in the article, and then randomly mask out some events for Event-Centric Pretraining. Notice only the training set is used for Pretraining. In the Task-Specific Contrastive Fine-tuning phase, we first replace all events with virtual tokens, whose initial representation is obtained by averaging the embedding of all tokens of the event. Finally, we input the mask embedding into an MLP classification layer to obtain the prediction probabilities for each candidate event.

\paragraph{\textsf{MCPredictor}}We first encode the event mentions and event texts, using the \textit{[CLS]} vector as the text representation. Then, we concatenate the two embeddings to obtain the initial representation $\textbf{e}_i$ of the event. Node that since the large number of candidate events, the computational complexity significantly increases when concatenating candidate events to the template to obtain event scores. So instead of employing \textit{Event-Level Scoring} and \textit{Script-Level Scoring}, We employ an MLP classifier on the template embeddings to obtain scores for each candidate event.

\par
All baseline experiments are conducted on PyTorch framework with CUDA on NVIDIA GTX 3090 Ti GPUs. We employ the RoBERTa-base\citep{Liu:et.al:2019:arXiv} for the base model, and set the length of sequence to 200, the mini-batch to 1, the training epoch to 15 on CGEP-ESC, while CGEP-MAVEN is 10.

\section{Details about Experimental Settings} \label{App:ExpSet}
Our method is implemented based on the pre-trained RoBERTa-base model~\cite{Liu:et.al:2019:arXiv} with 768-dimension provided by HuggingFace transformers~\footnote{https://github.com/huggingface/transformers}, and run PyTorch framework with CUDA on NVIDIA GTX 3090 GPU. We set the learning rate $l_{tr}$ for the PLM to $5e-6$, the weight coefficient $\beta$ to 0.5, and the temperature $\tau$ to 1.
\par
We use the average MRR (Mean Reciprocal Rank) and Hit@n (Hit Rate at n) overall impressions as the evaluation metrics, which are widely used in prediction and retrieval tasks~\cite{Chepurova:et.al:2023:EMNLP}.
For the \textit{HIT@n} metric, given an input sample $(G,n_c)$, if the model's top n predictions include $n_c$, then the model's prediction is deemed correct. Then the \textit{HIT@n} calculation formula is as follows:
$$\textit{HIT@n}=\frac{1}{N}\sum_{i=0}^{N} \mathbb{I}(Rank_i\leq n)$$
where $N$ means the total number of the data. For the \textit{MRR} metric, denote the predicted ranking of $n_c$ of the i-th sample among the candidate events as $Rank_i$, then the calculation formula for the metric is as follows:
$$\textit{MRR}=\frac{1}{N}\sum_{i=0}^{N} \frac{1}{Rank_i}$$
In this task, we employ \textit{MRR}, \textit{HIT@1}, \textit{HIT@3}, \textit{HIT@10}, and \textit{HIT@50} to measure the excellence of a model.

\begin{table*}[ht]
	\centering
	\resizebox{0.9\linewidth}{!}{
		\renewcommand{\arraystretch}{1.0}
		\begin{tabular}{clclccclllcllc}
			\toprule
			\midrule &  & \multicolumn{11}{c}{\cellcolor[HTML]{CBCEFB}\textbf{The Length of Event Chain}} & \\ \cmidrule{3-13}
			\multirow{-2}{*}{\textbf{MAVEN-ERE}} & \multirow{-2}{*}{\textbf{Docs}} & 3 & 4 & 5 & 6 & 7 & 8 & 9 & 10 & 11 & 12 & 13 & \multirow{-2}{*}{\textbf{Sum}} \\ \cmidrule{1-13}
			\textbf{Train} & 1552 & 2660 & 739 & 233 & 91 & 41  & 16 & 8 & 5 & 2 & 1 & 1 & \cellcolor[HTML]{FFCE93}\textbf{3797} \\
			\textbf{Valid} & 258 & 454 & 119 & 43 & 19 & 7 & 2 & 2 & 1 & 1 & 1 & 2 & \cellcolor[HTML]{FFCE93}\textbf{651}  \\
			\textbf{Test} & 258 & 454 & 140 & 50 & 20 & 5 & 1 & 0 & 0 & 0 & 0 & 0 & \cellcolor[HTML]{FFCE93}\textbf{670}  \\ \midrule
			\textbf{Sum} & \cellcolor[HTML]{FFCE93}\textbf{2068} & \cellcolor[HTML]{FFCE93}\textbf{3568} & \cellcolor[HTML]{FFCE93}\textbf{998} & \cellcolor[HTML]{FFCE93}\textbf{326} & \cellcolor[HTML]{FFCE93}\textbf{130} & \cellcolor[HTML]{FFCE93}\textbf{53} & \cellcolor[HTML]{FFCE93}\textbf{19} & \cellcolor[HTML]{FFCE93}\textbf{10} & \cellcolor[HTML]{FFCE93}\textbf{6} & \cellcolor[HTML]{FFCE93}\textbf{3} & \cellcolor[HTML]{FFCE93}\textbf{2} & \cellcolor[HTML]{FFCE93}\textbf{3} & \cellcolor[HTML]{FFCE93}\textbf{5118} \\
			\midrule
			\bottomrule
	\end{tabular}}
	\caption{Specific statistics of different division sets of the CGEP-MAVEN dataset for the \textit{script event prediction}.}
	\label{Tab:ScriptStatisticSplit}
\end{table*}

\begin{figure}[h]
	\centerline{\includegraphics[width=0.99\columnwidth]{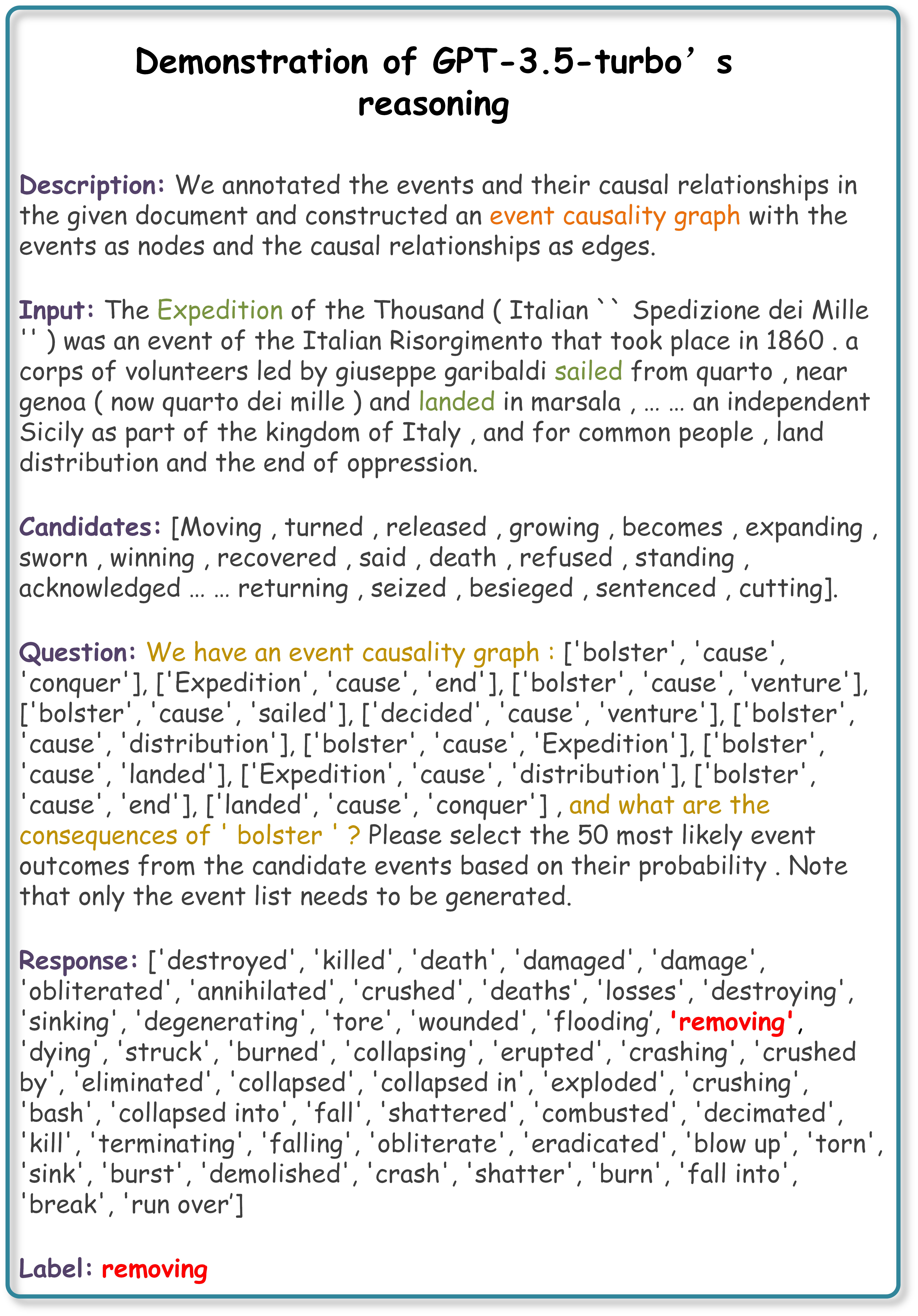}}
	\caption{A demonstration of GPT-3.5-turbo reasoning CGEP task.}
	\label{Fig:llmcase}
\end{figure}

\section{GPT-3.5-Turbo Prompt Detail} \label{GPT}
We evaluate GPT-3.5-turbo performance on the CGEP task under zero-shot settings. Figure~\ref{Fig:llmcase} illustrate a demonstration of GPT-3.5-turbo reasoning process. We first provide a formal definition of the causal event graph, then sequentially concatenate the text of each event in the event graph according to their indices to obtain the textual input. Simultaneously, we include the candidate events as input and linearize the event graph based on the weights of triplet. Finally, the model is queried with: "What are the subsequent events of {\textit{Anchor Event}}?" and then asked to select the top 50 most likely events from the candidate set.
\par
Note that when calculating the metrics, we remove events from the generated list that are not in the candidate set before computing the Hit@n metric. Additionally, when calculating MRR, if the golden event is not in the generated list, we uniformly assign it a rank of 256/512. Therefore, when generating the list, we typically allow it to generate more than 50 events, e.g. 60.

\section{The Datasets Process of SEP task}\label{Appendix:SEPDataset}
In this section, we will provide a detailed description of the data preprocessing for the \textit{script event prediction} task. For each CGEP-MAVEN instance, which contains an event causality graph, an anchor event, a candidate event set, and aground truth consequential event, we extract the longest event causality chains from the graph that terminate at the respective consequential nodes, and the other property, such as the candidate event set, the anchor event, will be maintained from the instance.
\par
Note that for each data instance, the maximum number of event chains extracted is limited to 1. Additionally, we remove any event chains containing fewer than 2 nodes. Ultimately, we divide the documents into training, validation, and test sets, with a split ratio of 75\%, 12.5\%, and 12.5\%, respectively. Table~\ref{Tab:ScriptStatisticSplit} summarizes the statics of final processed dataset for the task.

\end{document}